\documentclass[final]{cvpr}

\usepackage{times}
\usepackage{epsfig}
\usepackage{graphicx}
\usepackage{amsmath}
\usepackage{amssymb}
\usepackage{overpic}
\newcommand{\ys}[1]{{\color{blue}#1}}

\usepackage{mathtools}
\usepackage{amsmath}

\usepackage{amssymb}
\usepackage{amsfonts}
\usepackage{amsopn}
\usepackage{graphicx} % Necessary to use \scalebox
\usepackage{textcomp}
\usepackage{xfrac}
\usepackage{bbm}
\usepackage{overpic}
\usepackage{subfig}
\usepackage{wrapfig}
\usepackage[ruled,vlined,linesnumbered]{algorithm2e}
\usepackage{multirow}

\usepackage{color}
\definecolor{green}{rgb}{0, 0.5, 0}
\definecolor{orange}{rgb}{0.8, 0.6, 0.2}
\definecolor{red}{rgb}{1.0, 0.0, 0.0}
\definecolor{teal}{rgb}{0.0, 0.4, 0.4}
\definecolor{purple}{rgb}{0.65,0,0.65}
\definecolor{saffron}{rgb}{0.95,0.75,0.2}
\definecolor{turquoise}{rgb}{0.0,0.5,0.5}
\definecolor{brown}{rgb}{0.5, 0.16, 0.16}

\usepackage{overpic}
\usepackage{currfile} % \currfiledir

\newlength\savedwidth

\newcommand{\supl}[1]{{\color{black}\emph{#1}}}

\definecolor{lightgray}{rgb}{0.6, 0.6, 0.6}

\newcommand{\Fig}[1]{Figure~\ref{fig:#1}}
\newcommand{\Eq}[1]{Eq.~(\ref{eq:#1})}

\newcommand{\Sec}[1]{Section~\ref{sec:#1}}

\usepackage[normalem]{ulem}

\newcommand{\textapprx}{\raisebox{0.1ex}{\texttildelow}}

\newcommand{\hidecomment}[1]{}

\newcommand{\bR}{\mathbf{R}}
\newcommand{\bt}{\mathbf{t}}
\newcommand{\bT}{\mathbf{T}}
\newcommand{\bn}{\mathbf{n}}
\newcommand{\ba}{\mathbf{a}}
\newcommand{\bx}{\mathbf{x}}
\newcommand{\bI}{\mathbf{I}}
\newcommand{\bX}{\mathbf{X}}
\newcommand{\bv}{\mathbf{v}}
\newcommand{\br}{\mathbf{r}}

\newcommand{\bc}{\mathbf{c}}

\usepackage{xspace}
\usepackage[pagebackref=true,breaklinks=true,colorlinks,bookmarks=false]{hyperref}

\def\cvprPaperID{1205} % *** Enter the CVPR Paper ID here
\def\confYear{CVPR 2021}

\usepackage{cleveref}
\usepackage{tabularx}
\usepackage{nicefrac}
\usepackage{booktabs}

\begin{document}

%%%%%%%%% TITLE
%\title{StablePatches: Learning Object Poses from Geometrically Stable Patches\\of a Depth Image}
\title{StablePose: Learning 6D Object Poses from Geometrically Stable Patches}

\author{
    \hfill
	Yifei Shi\thanks{Joint first authors}\qquad
	Junwen Huang\footnotemark[1]\qquad
	Xin Xu\qquad
    Yifan Zhang\qquad
	Kai Xu\thanks{Corresponding author: kevin.kai.xu@gmail.com}
	\hfill
	\vspace{0.1cm}
	\\
	\hfill
	National University of Defense Technology
    \hfill
	%\vspace{-1cm}
}

\maketitle

%%%%%%%%% ABSTRACT
\begin{abstract}\vspace{-10pt}
   We introduce the concept of geometric stability to the problem of 6D object pose estimation and propose to learn pose inference based on geometrically stable patches extracted from observed 3D point clouds. According to the theory of geometric stability analysis, a minimal set of three planar/cylindrical patches are geometrically stable and determine the full 6DoFs of the object pose. We train a deep neural network to regress 6D object pose based on geometrically stable patch groups via learning both intra-patch geometric features and inter-patch contextual features. A subnetwork is jointly trained to predict per-patch poses. This auxiliary task is a relaxation of the group pose prediction: A single patch cannot determine the full 6DoFs but is able to improve pose accuracy in its corresponding DoFs. Working with patch groups makes our method generalize well for random occlusion and unseen instances. The method is easily amenable to resolve symmetry ambiguities. Our method achieves the state-of-the-art results on public benchmarks compared not only to depth-only but also to RGBD methods. It also performs well in category-level pose estimation.
   %Planar and cylindrical patches are ubiquitous in man-made objects. In this paper, we present a novel and general deep learning-based framework which predicts the 6D object pose by considering the relations of the planar and cylindrical patches. The key idea is to consider each patch locks only some of the DoFs of the pose, according to its slippage characteristics, and to predict the pose by the sampled geometrically stable patch triplets which generated by a stability analysis. The proposed method, which takes only the depth image as input, is especially applicable to 6D pose estimation of objects with occlusion. To our knowledge, this is the first method that learns 6D object pose by geometric stability analysis on patch. Moreover, we propose two network output representations to handle objects with discrete and continuous symmetries, respectively, which decrease the symmetry ambiguity. Experiments on multiple datasets demonstrate that the proposed method outperforms state-of-the-art methods by a large margin. Comprehensive studies demonstrate the effectiveness of our method on handling object symmetry, object occlusion and novel objects.
\end{abstract}\vspace{-20pt}

%%%%%%%%% BODY TEXT
%!TEX root = sceneparse.tex

\section{Introduction}
\label{sec:intro}

The problem of object pose estimation is to determine the 6D rigid transformation from the local object coordinate system to the camera reference frame.
Robust and accurate object pose estimation is of primary importance in a variety of applications ranging from robotic manipulation and localization to augmented reality.
Recent advances either predict correspondences between observations and template models~\cite{rad2017bb8}, or regress pose directly~\cite{xiang2017posecnn}.
In these tasks,
RGB features learned with convolutional neural networks have been predominantly adopted with notable success~\cite{hodan2020epos}.

%Deep features learned by convolutional neural networks from color information have been the dominant features to be used for inferring object poses.
Object pose inference with only color information, however, find difficulty in handling texture-less objects or unseen surface texture/appearance. In human perception, object pose hinges on object geometry~\cite{tarr1998three}. Humans cognize shapes and their poses simultaneously and in a coupled way~\cite{farah1988mental} in order to achieve a so-called invariant object recognition~\cite{karimi2017invariant}. Geometry enables a natural and powerful pose perception not only mitigating the distraction of color and appearance but also facilitating generalization to random occlusion and unseen instances.

3D geometric information does have been utilized for pose inference, both in traditional~\cite{drost2010model,vidal2018method} and in learning-based methods~\cite{zeng20173dmatch,chen2020g2l}, especially with the proliferation of depth cameras.
The most straightforward use is to perform ICP-based pose refinement with the geometric information~\cite{sundermeyer2018implicit,park2019pix2pose}.
Most deep learning approaches learn depth features to enhance color features~\cite{wang2019densefusion}.
Some others infer object poses from geometric features learned on 3D point clouds~\cite{gao20206d} or voxels~\cite{wada2020morefusion}.
These geometric features, however, are learned without an explicit guidance on the correlation between shape and pose, making pose reasoning based on them lack of interpretability and weak in generality.

%A minimal characteristic geometric features could pin down the pose of an object.

%!TEX root = ../sceneparse.tex

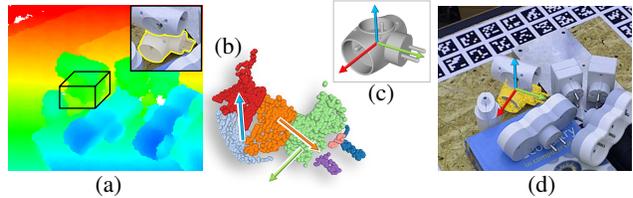
\begin{figure}
   \begin{overpic}[width=1.0\linewidth,tics=10]{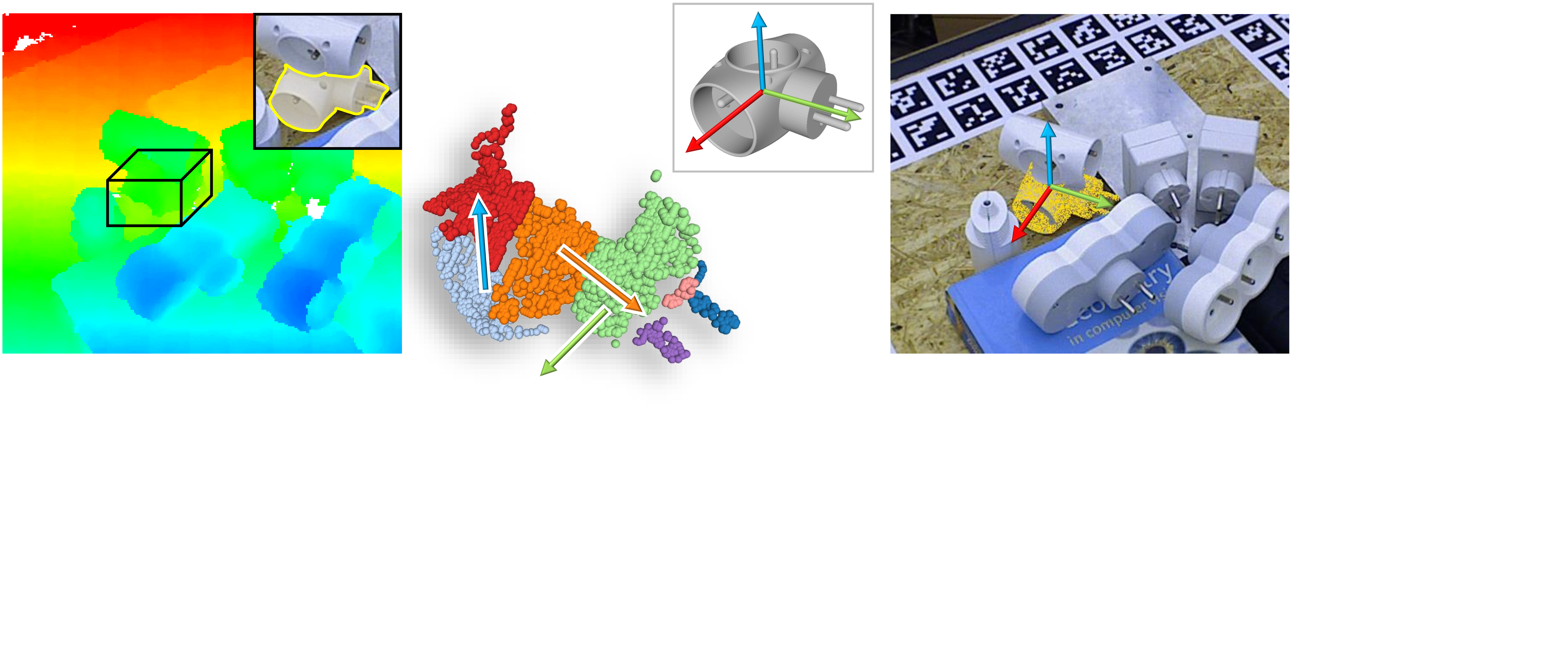}
   \put(14,-0.5){\small (a)}
   \put(33,21.5){\small (b)}
   \put(57.5,14.1){\small (c)}
   \put(83,-0.5){\small (d)}
   \end{overpic}
   \caption{Given the 3D point cloud (b) of a detected object (a), StablePose is trained to predict its 6D pose based on a geometrically stable patch group containing the blue, the orange and the green patches. Each patch determines a few DoFs and they altogether pin down all six DoFs (c, d).}
   \label{fig:teaser}\vspace{-12pt}
\end{figure} 

We propose to learn object pose inference based on \emph{3D surface patches} extracted from the point cloud of a single-view depth image.
In particular, we focus on planar and cylindrical patches which are omnipresent on the surface of household objects.
This design choice stems from two key insights. \emph{First}, patches are neither too local to capture meaningful geometric information, nor too global to be repeatable and generalizable across object instances.
\emph{Second}, each patch determines a specific set of DoFs of object pose. A minimal set of \emph{geometrically stable patches} can lock all six DoFs according to the theory of geometric stability (or slippage) analysis~\cite{gelfand2004shape}. It is therefore possible to accurately reason about 6D object poses over a small group of geometrically stable patches (\Fig{teaser}). Each stable group usually contains up to three patches, which facilitates fast learning. This also enables pose prediction with a redundant set of the stable groups, leading to robust pose estimation generalizing under occlusion and to unseen objects.

We design StablePose, a deep neural network trained to regress 6D object pose based on geometrically stable patch groups.
Given a patch-sample 3D point cloud, the network extracts both intra-patch geometric features and inter-patch contextual features. It then predicts a 6D pose for each stable patch group through aggregating the intra- and inter-patch features.
A dense \emph{point-to-point} pose loss is used to train this network.
A crucial design of StablePose is that a subnetwork is trained to predict per-patch poses.
This auxiliary task is a relaxation of the group pose prediction since a single patch cannot determine the full 6DoFs and thus the pose loss downgrades to a weaker \emph{point-to-patch} loss. Imposing such weak constraint for each patch in a stable group individually improves pose accuracy in the respective DoFs and altogether reinforces the group pose constraint, similar in spirit to the principle of geometrically stable ICP~\cite{gelfand2003geometrically}.
%Further, these weak tasks decouple the learning into different DoFs, making the network easier to train with faster convergence.

Given a 3D point cloud, we first extract a set of geometrically stable patch groups via performing stability analysis.
We then use StablePose to predict a 6D pose for each stable group. The final object pose takes the average of all group poses weighted by group stability.
To resolve ambiguities introduced by symmetries and achieve high pose accuracy, StablePose is trained to handle asymmetric objects and objects with discrete and continuous symmetries separately.
%In case of discrete symmetry, the we propose a new training scheme to back-propagate losses for multiple symmetries based on optimal assignment. Continuous rotational symmetry is handled by first regressing rotational axis and then computing 6D pose based on the predicted axis.
%
Through extensive evaluation, we show that StablePose outperforms state-of-the-art learning-based methods
by $19.1\%$ for depth-only input and $9.1\%$ for RGBD input
on the \texttt{T-LESS} benchmark~\cite{hodan2017t}.
%achieves the state-of-the-art performance among all depth-only methods and even beats the top-performing RGBD methods on several public datasets.
Furthermore, our method generalizes well for category-level pose estimation, obtaining competitive results on the \texttt{NOCS-REAL275} benchmark~\cite{wang2019normalized} and $18.6\%$ improvement on a more challenging ShapeNet based dataset.
Our work makes the following contributions:\vspace{-5pt}
\begin{itemize}
\item We, for the first time, introduce the concept of geometric stability into 6D object pose estimation.\vspace{-5pt}
\item We propose a deep network learning to infer 6D object pose based on geometrically stable patch groups. It attains high accuracy and robustness with a weak task of patch-wise, under-determined pose estimation.\vspace{-5pt}
\item We devise several key designs to accommodate a broad range of cases encompassing asymmetric or symmetric objects, objects with occlusion and unseen objects.
  %\item We achieve the state-of-the-art results on the T-LESS, YCB-Video and ShapeNet datasets, and demonstrate our method is generalize to novel objects by extensive experiments.
\end{itemize}

\section{Related work}
\label{sec:related}

\paragraph{Pose estimation from RGB}
The most common solution to object pose estimation from RGB images is to detect and match keypoints and solve a PnP. This approach has been well studied with a huge body of learned or non-learned methods (e.g.,~\cite{rad2017bb8,tekin2018real,peng2019pvnet,zakharov2019dpod,song2020hybridpose} and a survey~\cite{du2019vision}).
%due to its reliability on textured objects and under good lighting condition.
They are, however, less capable in the texture-less case where keypoints are hard to detect.

More recent works focus on predicting 6D object pose directly with trained deep neural networks~\cite{xiang2017posecnn}.
%Xiang et al.~\cite{xiang2017posecnn} proposed to regress 6D pose with a convolutional neural network.
Li et al.~\cite{li2018deepim} proposed a method which matches the rendered images of the object model against the input image for pose refinement. Similar idea was later explored in many follow-up works~\cite{labbe2020cosypose,wang2020self6d,park2020neural}.
SSD-6D~\cite{kehl2017ssd} and Deep-6DPose~\cite{do2018deep} integrate object detection, segmentation and pose regression from single-view RGB images in a unified network.
%These methods produce promising results in the case of moderate object occlusions.
Instead of directly regressing poses, another line of works learns to output interim results of pixel-wise 3D coordinates~\cite{brachmann2014learning,wang2019normalized,hodan2020epos,hu2020single}, based on which object poses can then be recovered.
Note, however, EPOS~\cite{hodan2020epos} samples 3D patches from \emph{template models} which are used as spatial clusters for 3D coordinates regression of image pixels. In contrast, our patches are sampled from \emph{observation point clouds} and serve as DoF determinant of 6D object pose.

%Learning-based pose estimation is often coupled with semantic segmentation or object detection, as the two tasks are highly related to each other.
%<PoseCNN><SSD-6D><A Unified Framework for Multi-View Multi-Class Object Pose Estimation><Deep Learning of Local RGB-D Patches for 3D Object Detection and 6D Pose Estimation><Implicit 3D Orientation Learning for 6D Object Detection from RGB Images><PVNet>

%Y. Xiang,W. Choi, Y. Lin, and S. Savarese, “Data-driven 3d voxel patterns for object category recognition,” in Proceedings of the IEEE Computer Vision and Pattern Recognition(cvpr), 2015, pp. 1903–1911.

%A. Mousavian, D. Anguelov, J. Flynn, and J. Kosecka, “3d bounding box estimation using deep learning and geometry,” in Proceedings of the IEEE Computer Vision and Pattern Recognition (cvpr), 2017.

%M. Sundermeyer, Z.-C. Marton, M. Durner, M. Brucker, and R. Triebel, “Implicit 3d orientation learning for 6d object detection from rgb images,” in European conference on computer vision, Springer, 2018, pp. 712–729.

%Self6D: Self-Supervised Monocular 6D Object Pose Estimation
\vspace{-10pt}
\paragraph{Pose estimation with RGB-D}
%With the proliferation of RGBD sensors, depth images start to play an important role in object detection and pose estimation.
The most straightforward use of depth is to perform pose refinement with ICP-based geometric alignment~\cite{sundermeyer2018implicit,park2019pix2pose} or by congruent set based registration~\cite{mitash2018robust}.
A more sophisticated approach is to utilize depth for 2D-3D feature fusion~\cite{wang2019densefusion}.
%propose to extract and fuse pixel-wise RGB features and point features for dense object pose prediction.
3D-keypoint-based approaches compute object poses by solving a least-square optimization to match detected 3D key-points and their counterparts of the template model~\cite{suwajanakorn2018discovery,he2020pvn3d}.
Point pair feature (PPF) based methods~\cite{drost2010model,vidal2018method,xu2019w} achieve high accuracy at the cost of high computation complexity.

The most relevant work to ours is~\cite{kehl2016deep} where a convolutional auto-encoder is trained to regress descriptors of locally-sampled RGB-D patches for 6D vote casting. During testing, scene patch descriptors are matched against model view patches and cast 6D object votes which are then filtered to refined hypotheses. While their method learns \emph{patch matching}, our method tries to learn \emph{pose regression} from geometrically stable patch groups.

\if 0
\paragraph{3D patch matching}
Patch matching has proven to be useful in a variety of applications, ranging from tracking and mapping~\cite{concha2015dpptam}, pose estimation~\cite{kehl2016deep}, segmentation~\cite{landrieu2018large} to reconstruction~\cite{hsiao2017keyframe,shi2018planematch}.
Previous works employed data-driven approaches to learn patch features for matching on RGB images~\cite{hodan2020epos}, 3D volumes~\cite{zeng20173dmatch} and point clouds~\cite{deng2018ppfnet}.
Our method does \emph{not} explicitly learn patch matching although the patch-wise pose prediction implicitly captures the patch-level correspondence between the observed point cloud and the template model. By stipulating that each patch predicts the ground-truth 6D pose in an under-determined manner, we do not require the supervision of per-patch matching other than a global pose.
%For patches on texture-less objects, the matching could be difficult to be learned as both the appearance and geometry could be indiscriminate.
%Shi et al.~\cite{} proposed a neural network that predicted the coplanarity between planar patches by utilizing the context in the RGB-D images. The approach is only capable of matching texture-less planar patches, such as wall and floor. In contrast, our method could match both planar patches and cylindrical patches.
\fi

\vspace{-10pt}
\paragraph{Stability analysis}
Stability or slippage analysis is a powerful tool of shape analysis.
It is originally posed to subsample points from a point set while maintaining the stability of DoFs in ICP-based alignment~\cite{gelfand2003geometrically,pottmann2003geometry,brown2007global}.
This is done by filtering out redundant points while keeping sufficient points for each alignment DoF.
Another application of stability analysis is to extract slippage signatures for discovering slippable components on a 3D object~\cite{gelfand2004shape,bokeloh2008slippage}.
Inspired by these works,
%In contrast, we introduce to use the stability analysis in 6D object pose estimation from single-view depth images.
our work, for the first time, introduces the concept of stability into object pose estimation.

%~\ys{
%\subsection{Handling object symmetry}
%} 
%!TEX root = sceneparse.tex

\section{Stability Analysis for Pose Estimation}
\label{sec:review}

\begin{figure}
   \begin{overpic}[width=1.0\linewidth,tics=10]{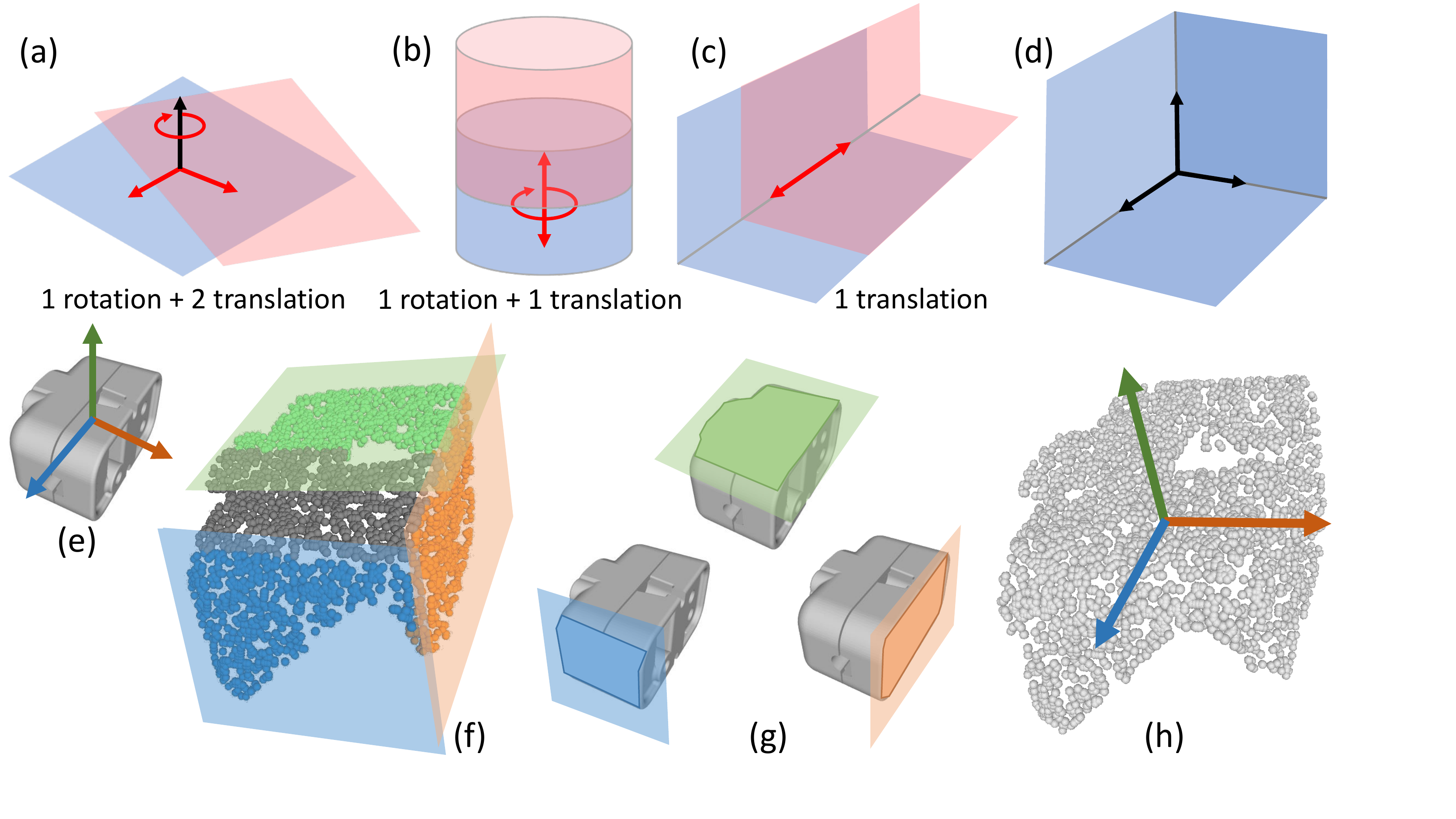}
   \end{overpic}
   \caption{(a-d): Geometric stability of different shapes. The unstable/slippable transformations of each shape are annotated with red arrows. A group of three non-coplanar planes (d) is geometrically stable under rigid transformation. (e-h): Using geometric stability for object pose estimation. Given the observation of an object (f), a geometrically stable patch group is extracted and matched to the model shape (g), thus determining the 6D pose of the object (h).}
   \label{fig:stability}\vspace{-14pt}
\end{figure}

We describe stability analysis of 3D shapes, also known as slippage analysis, and investigate its relation to object pose estimation. The geometric stability of a 3D shape can be characterized by the rigid transformations (a translation and a rotation) that minimizes point-to-plane error metric.
Intuitively, if a 3D shape is transformed by a rigid transformation without introducing significant motion along the normal direction at each surface point, it is called geometrically unstable (or slippable) along that transformation. If there is no such transformation, the shape is stable. See Figure~\ref{fig:stability}(a-d) for a few examples with different stability.

Given a 3D shape sampled into a point set, we can compute a stability measure base on the eigenvalues of the $6\times 6$ covariance matrix of the 6D rigid transformation minimizing point-to-plane error. \supl{Please refer to the appendix in the supplemental material for details of this computation.}

\vspace{-10pt}
\paragraph{Stability and object pose}
%\kx{We, for the first time, introduce the concept of stability for object pose estimation.}
Given the geometric observation of an object, we solve the problem of 6D object pose estimation on the basis of stability analysis. Specifically, we first extract a group of planar or cylindrical patches which are geometrically stable. The patches in the stable group are then aligned to the corresponding patches on the 3D model in canonical pose. Each alignment determines a subset of the six DoFs. All patches in the stable groups together pin down the 6D pose of the object. See Figure~\ref{fig:stability}(e-h) for an illustration.
In practice, however, finding patch correspondence itself is a challenging problem.
In what follows, we propose a deep neural network which predicts 6D object poses without relying on patch correspondences.

%!TEX root = ../sceneparse.tex

\begin{figure*}[t!] \centering
	\begin{overpic}[width=1.0\linewidth,tics=10]{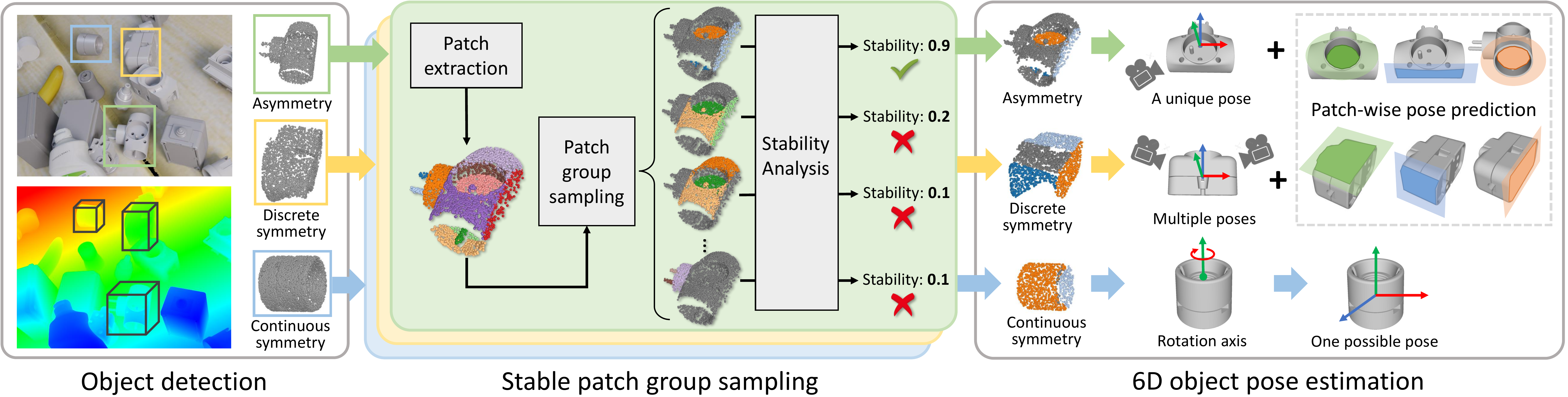}%,grid
   \end{overpic}\vspace{-2pt}
   \caption{Method overview. Given a single-view RGB-D image, we first detect and segment objects in the RGB image and then crop the depth point cloud. Based on object labels, the objects are categorized according to symmetry property. For each object point cloud, we extract planar and cylindrical patches and sample a set of geometrically stable patch groups based on stability analysis. Our network predicts for stable group a 6D pose, with patch-wise pose estimation as an auxiliary task. For objects with continuous symmetry, it computes the rotation axis and then outputs one possible 6D pose.
   }
   \label{fig:overview}\vspace{-12pt}
\end{figure*} 

\section{Method}
\label{sec:method}

\paragraph{Overview}
Figure~\ref{fig:overview} provides an overview of our method. The input to our method is an RGBD image capturing one or multiple objects. The RGB image is used only for object detection but \emph{not} for pose estimation. The output is the 6DoF pose of each object. Our method starts from detecting objects and predicting their 2D masks in the depth image. For each detected object, we obtain a 3D point cloud by unprojecting its depth mask. We then extract planar and cylindrical patches from the point cloud (\Sec{patchext}) and sample a set of geometrically stable patch groups based on stability analysis (\Sec{stability}). Each stable group is fed into a deep network to predict the 6D pose of the corresponding object and the final pose is obtained by averaging the 6D poses obtained with all stable groups (\Sec{poseest}). To resolve the ambiguities caused by symmetry, our network handles asymmetric objects and objects with discrete and continuous symmetries separately, each trained with a proper loss.

\subsection{Object Detection and Patch Extraction}
\label{sec:patchext}
The first step is to detect objects and extract patches. Any RGB-based object detection method can be used here. We adopt the detector proposed in Pix2Pose~\cite{park2019pix2pose} which is also used in many recent pose estimation works.
We then crop the depth image based on the detected object mask and unproject the cropped depth image into a 3D point cloud.
For each object point cloud, we extract both planar patches $\mathcal{P}^\text{P}=\{p_i^\text{P}\}$ and cylindrical patches $\mathcal{P}^\text{C}=\{p_i^\text{C}\}$. Let $\mathcal{P}=\mathcal{P}^\text{P} \cup \mathcal{P}^\text{C}$. Again, many existing methods can be utilized to extract patches from point clouds.
We found through experiment that CAPE~\cite{proencca2018fast} is fast and relatively robust for our data modality (i.e., point cloud converted from single-view depth image of objects with occlusion).

\subsection{Stable Patch Group Sampling}
\label{sec:stability}
One could predict object pose based on all patches.
%Instead of using all the patches to make the prediction, we predict pose on the sampled patches.
Ideally, the more patches are used, the more global information is encoded and the better the pose can be estimated~\cite{kehl2016deep}. In the case of single-view observations, however, using all patches may not be a good choice since the patch count varies from different views due to occlusion. Based on the fact that 6D object pose can be determined with a minimum of three geometrically stable patches (e.g., nearly mutual orthogonal planar patches in \Fig{stability}), we opt for learning pose estimation based on \emph{patch triplets}.

%According to \cite{shi2018planematch}, the minimal number of planar or cylindrical patches to make an alignment stable is $3$.
Given an object point cloud, we enumerate all triplets out of $\mathcal{P}$ as patch group candidates. We then analyze the stability of each group.
%In particular, we sample a $1000$ points for each patch group candidate and compute the covariance matrix $C$ with the points using \Eq{covariance}.
%\ys{The stability is measured as $[1+e^{0.05\left(\frac{\lambda_6}{\lambda_1}-200\right)}]^{-1}$, where $\lambda_1$ and $\lambda_6$ are the smallest and largest eigenvalues of $C$ respectively. }
%We collect those groups whose stability value is greater than $0.5$
We collect those groups whose stability measure passes the threshold
into the set of geometrically stable patch groups $\mathcal{G}=\{G_k=(p_{k1},p_{k2},p_{k3})\}$.
To enhance the generality of our pose estimation network, we train it with patch groups sampled from point clouds in multiple views for each object.

\subsection{Object Pose Estimation}
\label{sec:poseest}
%We first describe the networks for intra- and inter-patch feature extraction and then the networks for predicting 6D pose for asymmetric objects, followed by the handling of objects with discrete and continuous symmetries.

\subsubsection{Patch Feature Extraction}

\paragraph{Intra-patch geometric feature}
To learn geometric feature for each individual patch, we utlize PointNet++~\cite{qi2017pointnet++}. The input is the point cloud of a patch (resampled to $2000$ points) and the output is a feature vector of $1024$D. Note that this learned feature encodes for a patch not only its local geometry but also its relative position in the whole object. The latter benefits spatial reasoning in pose estimation.

\vspace{-10pt}
\paragraph{Inter-patch contextual feature}
To help our network reasoning about object pose globally, we additionally learn inter-patch contextual features using all patches.
%Since object pose is a non-local property, so the global feature should be involved in pose estimation.
Since we have extracted per-patch features, we concatenate the per-patch features for every two patches and then aggregate them using the Relation Network~\cite{santoro2017simple}. Relation network provides a mechanism for non-local message passing between patches. The final feature is the adaptive pooling of the pair-wise features after message passing, which encodes inter-patch structure and global shape context.

\subsubsection{Pose Estimation of Asymmetric Objects}

\begin{figure*}[t!] \centering
	\begin{overpic}[width=0.97\linewidth,tics=10]{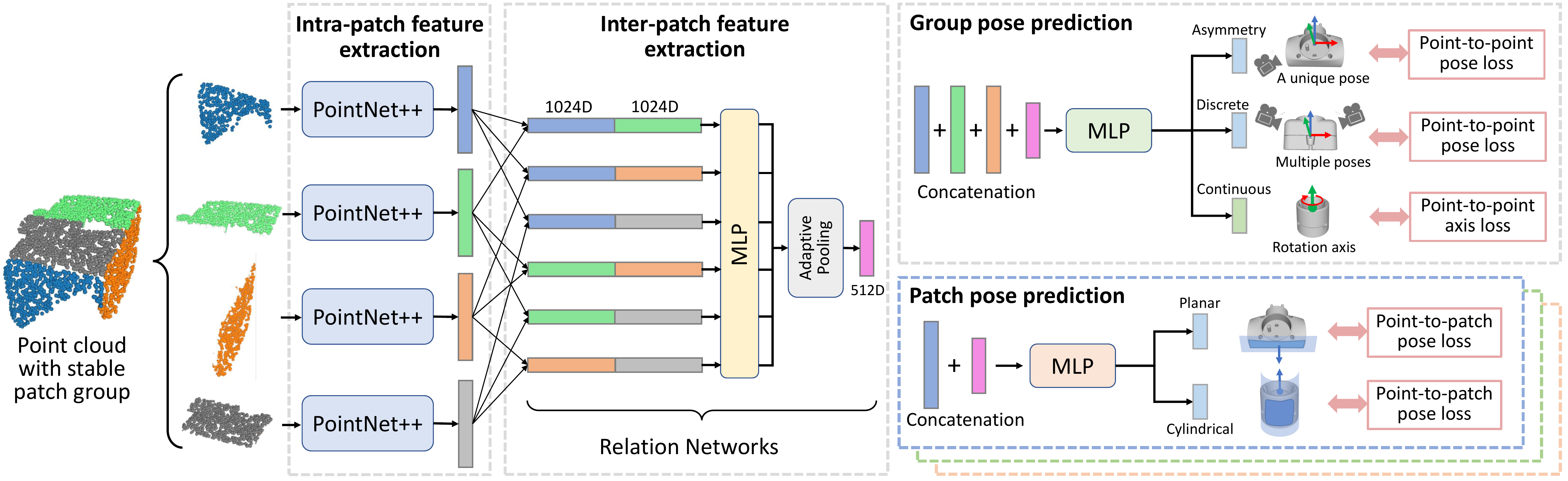}%,grid
   \end{overpic}\vspace{-5pt}
   \caption{The network architecture of StablePose which is composed of four subnetworks.}
   \label{fig:network}\vspace{-15pt}
\end{figure*}

\paragraph{Group pose prediction}
Given an object, its 6D pose is predicted on the basis of its geometrically stable patch
groups $\mathcal{G}$.
As shown in Figure~\ref{fig:network}, for each $G_k$ in $\mathcal{G}$, a three-layer multi-layer perceptrons (MLPs) takes the concatenation of its per-patch geometric features and the inter-patch contextual feature as input and outputs a 6D pose.
To train the network, we impose a \emph{dense-point pose loss} which measures the discrepancy between the predicted 6D pose and the ground-truth on a per-point basis:
\begin{equation}\label{eq:gpl}\small
%\mathcal{L}^\text{group}_k=\sum_{j}\|(\bR^\text{G}_{k}\bx_j+\bt^\text{G}_k)-(\bar{\bR}\bx_j + \bar{\bt})\|
\mathcal{L}^\text{group}_k=\sum_{j}\|\bT^\text{G}_{k}\bx^\text{m}_j - \bar{\bT}\bx^\text{m}_j\|,
\end{equation}
where $\bT^\text{G}_k=[\bR^\text{G}_k|\bt^\text{G}_k]$ is the predicted pose for group $G_k$ and $\bar{\bT}_i=[\bar{\bR}|\bar{\bt}]$ the ground-truth object pose.
$\mathcal{X}^\text{m}=\{\bx^\text{m}_j\}$ are the points sampled on the template 3D model.

%!TEX root = ../sceneparse.tex

\begin{figure}[b!]\vspace{-10pt}
   \begin{overpic}[width=1.0\linewidth,tics=10]{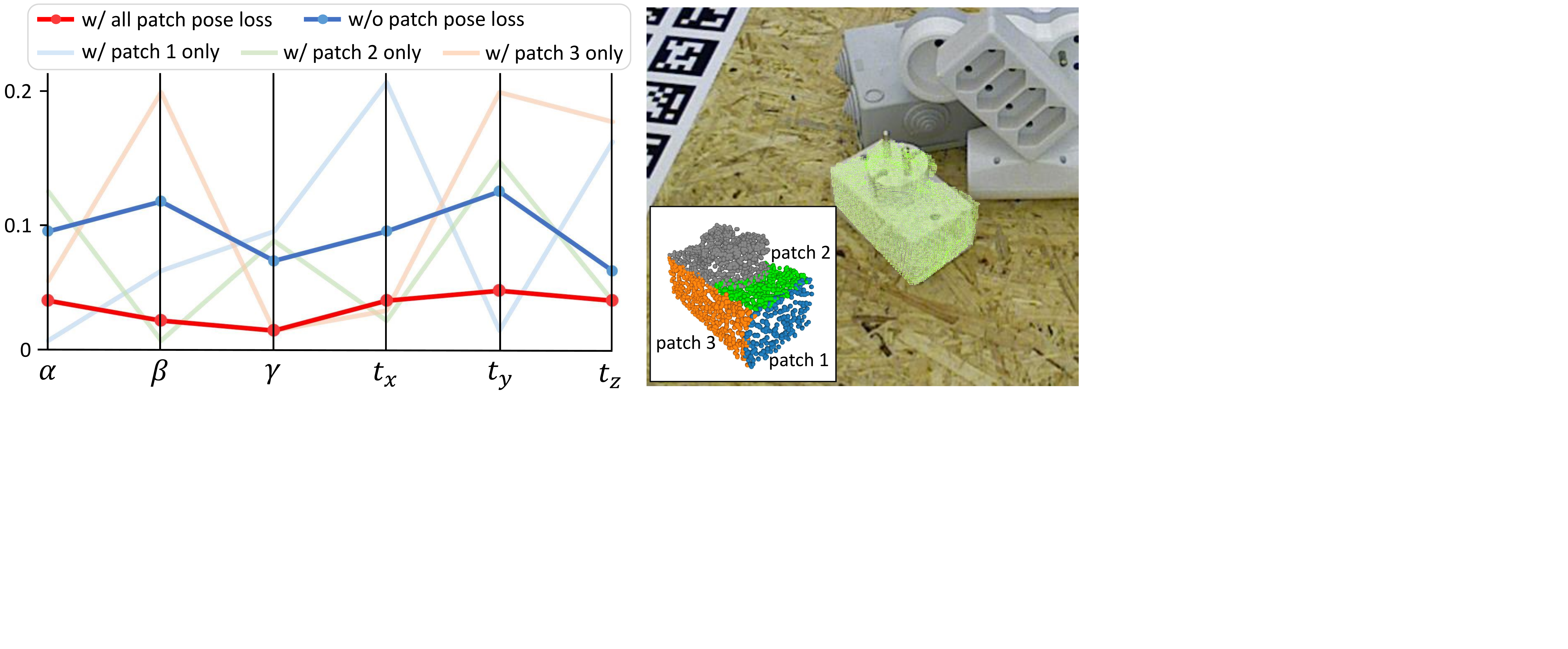}
   \end{overpic}\vspace{-2pt}
   \caption{A parallel coordinates visualization~\cite{inselberg1990parallel} of object pose errors in the six DoFs. Minimizing the pose loss of one patch optimizes a specific subset of DoFs (see the curves with light shading). Minimizing the group pose loss but not per-patch pose loss leads to high overall pose error (the blue curve). The best overall accuracy is obtained by minimizing both group and all-patch pose losses (the red curve).}
   \label{fig:patchloss}
\end{figure}

\vspace{-10pt}
\paragraph{Patch pose prediction}
In addition to the group pose prediction, we introduce an auxiliary task of patch-wise pose prediction.
This task is a relaxation of the group pose prediction since a single patch cannot determine all six DoFs and thus the point-to-point loss in \Eq{gpl} needs to be replaced by a weaker ``point-to-patch'' loss. For a stable patch group, imposing such weak constraint for each of its three patches individually improves the pose accuracy in respective DoFs and altogether reinforces the group pose constraint (see \Fig{patchloss}). This process resembles the stability-based ICP for 3D shape registration~\cite{shi2018planematch}. Further, these weak tasks decouple the learning into different sets of DoFs, making the network easier to train with faster convergence.

For each patch $p_i \in \mathcal{P}$, we devise a three-layer MLP which takes the concatenation of its geometric feature and the contextual feature as input and produces a patch pose $\bT^\text{P}_i=[\bR^\text{P}_i|\bt^\text{P}_i]$. The network is trained by minimizing a \emph{point-to-plane loss} for planar patches or a \emph{point-to-axis loss} for cylindrical patches.
In particular, we define the dense point-to-plane loss for planar patches as
\begin{equation}\label{eq:ppl}\small
%\mathcal{L}^\text{planar}_{ki}=\sum_{j}\|[\bR^{-1}_{k}(\bx_j-\bt_k)-\bar{\bR}^{-1}(\bc_i-\bar{\bt})]\cdot (\bar{\bR}^{-1}\bn_i)\|
\mathcal{L}^\text{planar}_{i}=\sum_{j}\|[(\bT^\text{P}_i)^{-1}\bx_{ij}-(\bar{\bT})^{-1}\bc_i] \cdot (\bar{\bR}^{-1}\bn_i)\|,
\end{equation}
where $\bc_i$ represents the center of patch $p_i$, and $\bn_i$ is the normal of the plane that $p_i$ lies in.
$\bx_{ij}$ is a point on $p_i$.
For cylindrical patches, point-to-patch error can be reduced to point-to-axis error and the corresponding patch pose loss is
\begin{equation}\label{eq:cpl}\small
%\mathcal{L}^\text{cylind}_{i}=\sum_{j}\|d((\bT^\text{P}_i)^{-1}\bx_j,\bar{\phi}_i)-d(\bar{\bT}^{-1}\bc_i,\bar{\phi}_i)\|,
\mathcal{L}^\text{cylind}_{i}=\sum_{j}\|d((\bT^\text{P}_i)^{-1}\bx_{ij},\bar{\phi}_i)-r_i\|,
\end{equation}
where $\bar{\phi}_i$ is axis of the patch cylinder transformed with $\bar{\bT}^{-1}$ and $r_i$ the radius of the cylinder.
$d(\bx, \phi)$ denotes the distance between point $\bx$ and line $\phi$.

\vspace{-10pt}
\paragraph{The overall loss}
The 6D pose loss for asymmetric objects sums up the pose losses over all stable patch groups and those of all planar and cylindrical patches:
\begin{equation}\label{eq:asymloss}\small
\mathcal{L}^\text{asym}=\sum_{k=1}^{|\mathcal{G}|}{\mathcal{L}^\text{group}_k}+\sum_{i=1}^{|\mathcal{P}^\text{P}|}{\mathcal{L}^\text{planar}_i}
+\sum_{i=1}^{|\mathcal{P}^\text{C}|}{\mathcal{L}^\text{cylind}_i}
\end{equation}

%!TEX root = ../sceneparse.tex

%A straightforward way to aggregate loss over all the predictions is: $\mathcal{L}=\sum_{i}{\mathcal{L}_i}$.
%However, the prediction made by each patch group are not equally accurate. To improve the performance during testing, we train the network to make it decide which predicted pose is likely to be the best in an unsupervised manner.
%This is achieved by modifying the network to output an extra confidence score $s_i$ corresponding to each prediction, in addition to the pose prediction.
%The loss function with a confidence regularization term is:
%\begin{equation}\label{eq:sumloss}
%\mathcal{L}=\frac{1}{N}\sum_{i}^{N}(\mathcal{L}_is_i-w\log s_i)
%\end{equation}
%where $w$ is a pre-defined balancing parameter. The regularization term aims at filtering the inaccurate and low confident predictions. An interesting observation is that, in the experiments, the learned confidence scores are largely corresponds to the stability of the patch groups, i.e. the patch groups with high stability owns high confidence scores.

\subsubsection{Pose Estimation of Symmetric Objects}

\paragraph{Discrete symmetry case}
Object symmetries introduce ambiguities to pose estimation.
A common strategy of resolving symmetry ambiguities in training a pose estimation network is ``back-propagate with the best''~\cite{wang2019densefusion,labbe2020cosypose}. In particular, suppose the object possesses a discrete set of symmetries $\{\mathbf{T}^\text{S}_i\}_{i=1,...,M}$
\footnote{A symmetry transformation $\mathbf{T}^\text{S} \in {\rm SE}(3)$ of a 3D shape $\bX$ satisfies $\bT^\text{S}\bX=\bT^\text{S}$ and $\bT^\text{S} \neq \bI$.}
and its 6D pose estimated by the current neural network is $\bT$. In the next round of training, we choose the transformation from $\{\bT, \bT\mathbf{T}^\text{S}_1, \ldots, \bT\mathbf{T}^\text{S}_M\}$ such that it leads to the minimal alignment error between the observation and the template model, and back-propagate this error.
Using this scheme, however, the network tends to memorize a particular pose of the object and lacks a global understanding of object symmetries resulting in weak generality on occluded or unseen objects. To overcome this issue, we propose a new training method for handling symmetry ambiguities --- ``\emph{back-propagate with all}''.

%For objects with discrete symmetries, let $\mathcal{T}^\text{S}=\{\mathbf{T}^\text{S}_i\}_{i=1,...,M}$ denote the set of symmetry transformations which is a rigid transformation mapping the shape to itself:
%\begin{equation}\label{eq:ref_sym}
%\mathcal{T}^\text{S} = \{\mathbf{T}^\text{S}\in {\rm SE}(3) | \bT^\text{S}\bX=\bT^\text{S} \text{ and } \bT^\text{S} \neq \bI \}
%\end{equation}

In our approach, instead of estimating only a single object pose, we predict $M+1$ object poses, each against one of the ground-truth poses in $\{\bar{\bT}, \bar{\bT}\mathbf{T}^\text{S}_1,\ldots,\bar{\bT}\mathbf{T}^\text{S}_M\}$, using again a three-layer MLP.
%This is achieving by using multiple MLPs, as shown in Fig?.
Since there are multiple outputs, the correspondence between the predictions and ground-truths should be determined on-the-fly, so that the loss of each individual prediction could be evaluated and back-propagated.
To this end, we devise an optimal assignment process~\cite{kuhn1955hungarian} which finds a benefit-maximizing correspondence by solving the following optimization:
\begin{equation}\label{eq:assignment}\small
\left.\begin{aligned}
  &\mathop{\arg\max}_{\Pi} \sum_{m=1}^{M+1}\sum_{k=1}^{M+1}B_{m,k}\Pi_{m,k},\nonumber\\
%  \text{s.t.} \ \sum_{m=1}^{M+1}&\Pi_{m,k}=1, \text{ } k\in\{1,\ldots , M+1\}; \\
%  \sum_{k=1}^{M+1}&\Pi_{m,k}=1, \text{ } m\in\{1,\ldots , M+1\}.
  \text{s.t.} \ \sum_{m=1}^{M+1}\Pi_{m,k}&=1,\sum_{k=1}^{M+1}\Pi_{m,k}=1,\text{ } m,k\in\{1,\ldots , M+1\}.\nonumber
\end{aligned}\right.
\end{equation}
$\mathbf{\Pi}$ is a permutation matrix with $\Pi_{m,k}\in\{0,1\}$ indicating whether the $m$-th predicted pose matches the $k$-th ground-truth pose. $M+1$ is the total number of possible object poses given $M$ object symmetries. $\mathbf{B}$ is a benefit matrix in which $B_{m,k}$ represents the benefit of matching the $m$-th predicted pose to the $k$-th ground-truth. A higher similarity in pose leads to a larger benefit.
The benefit $B_{m,k}$ between two poses $\bT_{m}$ and $\bT_{k}$ is computed as the point-wise Euclidean error between point sets transformed by the two poses respectively:
\begin{equation}\label{eq:benefit}\small
 B_{m,k}= \sum_{j} \| \bT_m \bx_j - \bT_k \bx_j \|,
\end{equation}
where $\mathbf{X}=\{\bx_j\}$ are sample points on the template model.

Once the correspondences are determined, the pose loss can be evaluated by accumulating the asymmetry loss in \Eq{asymloss} over all $M+1$ possible poses under $M$ symmetries:
\begin{equation}\label{eq:sumloss2}\small
%\mathcal{L}_\text{sym}=\sum_{m=1}^{M+1}(\sum_{i=1}^{|\mathcal{G}|}{\mathcal{L}^{pose}_i}+\sum_{i=1}^{|\mathbf{P}|}{\mathcal{L}^{cop}_i})
%\mathcal{L}^\text{dsym}=\sum_{\bT \in \mathcal{T}^\text{S}} \mathcal{L}^\text{asym} (\bT),
\mathcal{L}^\text{dsym}=\sum_{\bT^\text{S} \in \mathcal{T}^\text{S}} \mathcal{L}^\text{asym} (\bT\rightarrow\bT\bT^S).
\end{equation}
$\mathcal{L}^\text{asym} (\bT\rightarrow\bT\bT^S)$ means that the loss is computed by replacing the predicted group/patch pose $\bT$ with $\bT\bT^S$, with $\bT^S \in \mathcal{T}^\text{S}$ and $\mathcal{T}^\text{S} = \{\bI, \bT^S_1, \ldots, \bT^S_M$\}.

%Overall, the loss of discrete symmetric object is sum of the total pose loss by all the geometrically stable patch groups and the total coplanarity loss by all the planar patches.
%\begin{equation}\label{eq:discretesumloss}
%\mathcal{L}^\text{discrete symmetry}=\frac{1}{N}\frac{1}{M}\sum_{i}^{N}\sum_{m}^{M}(\mathcal{L}_{im}s_{im}-w\log s_{im})
%\end{equation}
\vspace{-10pt}
\paragraph{Continuous symmetry case}
For object with a continuous rotational symmetry, the number of ground-truth pose is infinite.
Therefore, instead of predicting poses directly, we opt to train a three-layer MLP to regress
the rotation axis represented by the object center $\mathbf{c}^\text{r}$ and the orientation vector of the axis $\mathbf{a}^\text{r}$, based on each stable group.
Let us define $\mathbf{T}(\theta,\bc^\text{r},\ba^\text{r})$ as the transformation of rotating an angle of $\theta$ about the axis $(\mathbf{c}^\text{r},\mathbf{a}^\text{r})$. The regression loss of rotation axis for each patch group $G_k$ is then defined as the point-wise Euclidean error between model point sets transformed by the predicted and ground-truth rotational transformations respectively:
\begin{equation}\label{eq:rotloss}\small
  \mathcal{L}^\text{rot}_k=\dfrac{1}{|\Theta|}\sum_{\theta \in \Theta}\sum_{j}\|\bT(\theta, \bc^\text{r},\ba^\text{r})\bx_j-\bT(\theta, \bar{\bc}^\text{r},\bar{\ba}^\text{r})\bx_j\|,
\end{equation}
%where $\mathbf{T}^{r}_\gamma$ is the rotational transformation of $\mathbf{T}^{r}$ with a rotation angle of $\gamma$.
where $(\bar{\bc}^\text{r},\bar{\ba}^\text{r})$ is the ground-truth rotation axis.
The loss is computed over a set of rotation angles $\Theta=\{\kappa\cdot\pi/8\}_{\kappa=1,\ldots,16}$.
The overall loss sums up the losses of both per-group axis prediction and per-patch pose prediction:
\begin{equation}\label{eq:symsumloss}\small
\mathcal{L}^\text{csym}=\sum_{k=1}^{|\mathcal{G}|}\mathcal{L}^\text{rot}_k+\sum_{i=1}^{|\mathcal{P}^\text{P}|}{\mathcal{L}^\text{planar}_i}
+\sum_{i=1}^{|\mathcal{P}^\text{C}|}{\mathcal{L}^\text{cylind}_i}.
\end{equation}
%The final rotation axis is computed by averaging the axes regressed by all stable groups.
Having obtained the rotation axis, we can compute the 6D object pose $[\bR|\bt]$.
The translation DoFs are determined: $\bt=\bc^\text{r}$.
The quaternion corresponding to $\bR$ is
$$
q^R = \left(\cos\frac{\gamma}{2}, \sin\frac{\gamma}{2}a_x, \sin\frac{\gamma}{2}a_y, \sin\frac{\gamma}{2}a_z\right)
$$
where $\ba^\text{r}=(a_x, a_y, a_z)$ is the axis vector and $\gamma$ can be an arbitrary angle (we set $\gamma=\frac{\pi}{2}$) since there is one rotational DoF that cannot be determined by the rotation axis.

\subsubsection{Inference}
During inference, the 6D pose for each stable patch group is first regressed with the proper network according to the object's symmetry property. The final object pose is the weighted average over all group poses, with the weight being the  stability measure of a group. Higher stability leads to higher weight. Note that the weighed average of rotation and translation are computed separately. The weighted combination of rotations is computed in quaternion form.

%!TEX root = sceneparse.tex

\subsection{Implementation details}
\label{sec:implementation}

%\paragraph{Network architecture}
%The inter-patch contextual feature is 512D.
The number of points sampled is $500$ on template models
%in Eq. (\ref{eq:alignerror}), (\ref{eq:benefit}) and (\ref{eq:rotloss}),
in Eq. (\ref{eq:gpl}), (\ref{eq:benefit}) and (\ref{eq:rotloss}),
and $100$ for each patch in Eq. (\ref{eq:ppl}) and (\ref{eq:cpl}).
The dimensions of several critical features are shown in \Fig{network}.
Rotations are represented in quaternion form.
%\paragraph{Training and inference}
%We implemented StablePose using PyTorch and trained it on an NVIDIA Tesla T4 GPU.
The network is trained first with patch-wise pose losses only for $2\textapprx3$ epoches as a warmup and then with all losses turned on. %This training strategy leads to fast convergence.
The batch size is $16$.
%We use the ADAM optimizer with a learning rate of $0.0001$.
For some small objects with less than three patches extracted, we use PointNet++~\cite{qi2017pointnet++} to extract feature for the object points and then regress the pose/axis directly with a three-layer MLP.
%The mechanism to handing symmetry is also utilized.
For objects with no stable patch group extracted, we simply use the inter-patch contextual feature to make pose/axis prediction. The ratio of these two cases is $10\textapprx30\%$ for the datasets we have tested.

%!TEX root = sceneparse.tex

\newcommand{\tless}{\texttt{T-LESS}\xspace}
\newcommand{\lmo}{\texttt{LineMOD-O}\xspace}
\newcommand{\nocs}{\texttt{NOCS-REAL275}\xspace}
\newcommand{\snp}{\texttt{ShapeNetPose}\xspace}
\newcommand{\eadi}{$e_\text{ADI}$\xspace}
\newcommand{\eadiadd}{$e_\text{ADI/ADD}$\xspace}
\newcommand{\evsd}{$e_\text{VSD}$\xspace}

\section{Results and Evaluation}
\label{sec:results}

%!TEX root = ../sceneparse.tex

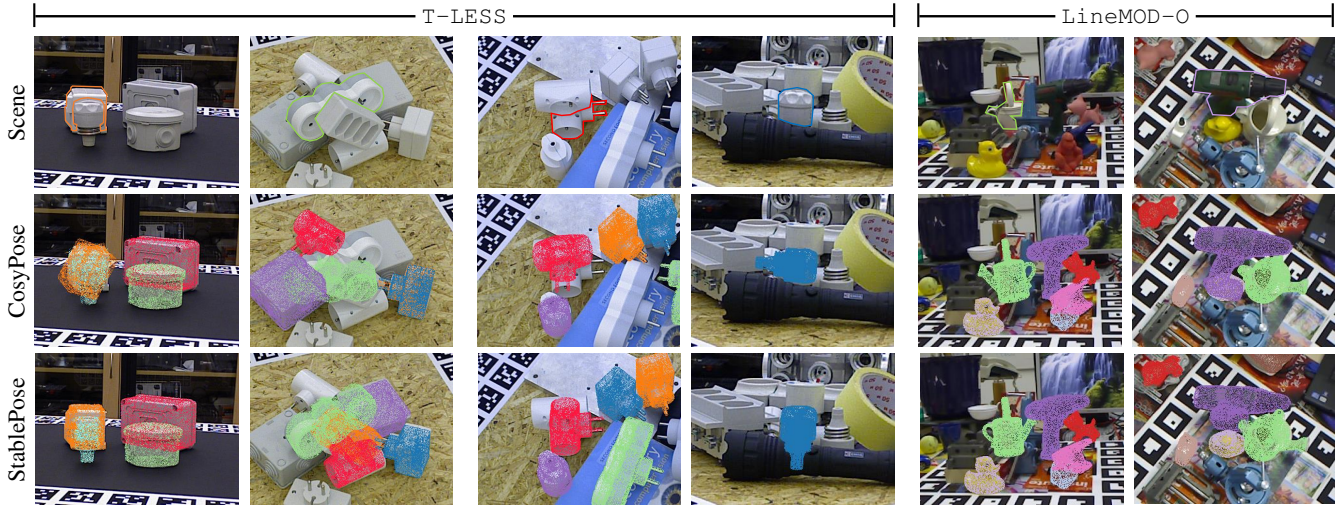
\begin{figure*}[t!] \centering
	\begin{overpic}[width=1.0\linewidth,tics=5]{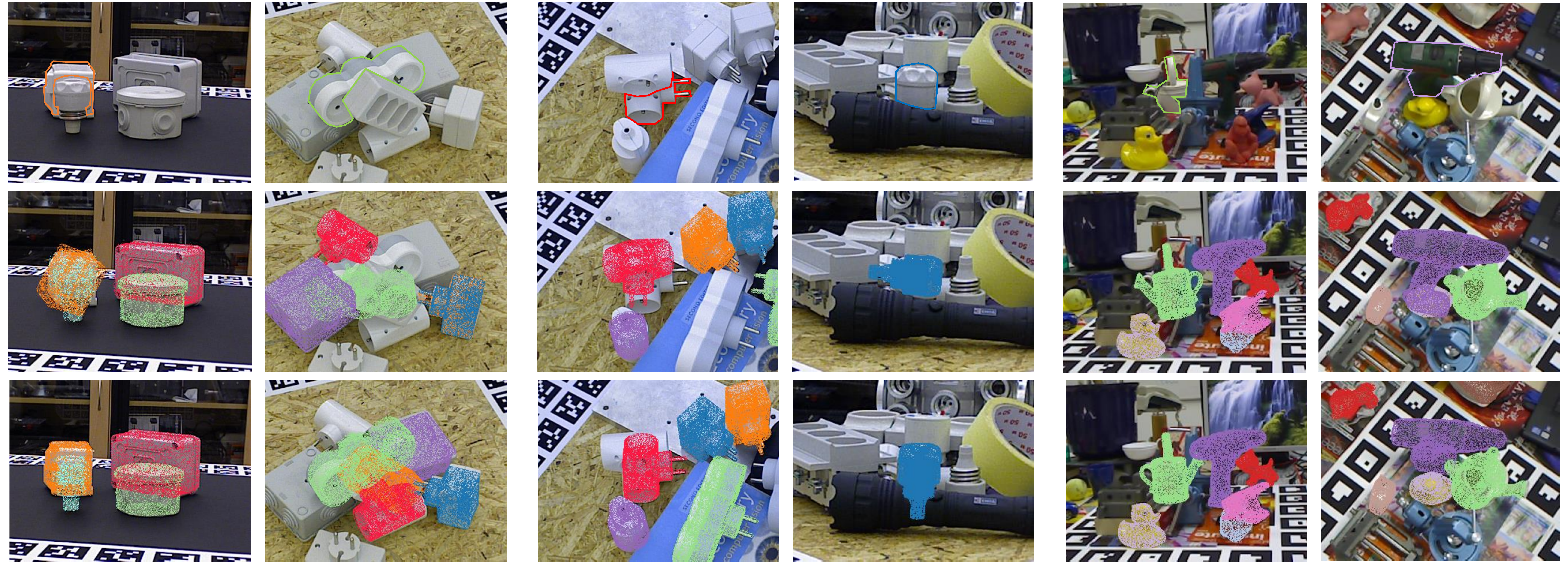}%,grid
    \put(-1.5,2){\rotatebox{90}{\small StablePose}}
    \put(-1.5,14.5){\rotatebox{90}{\small CosyPose}}
    \put(-1.5,28){\rotatebox{90}{\small Scene}}
    \put(0.4,37){\small \rule[-0.4ex]{0.2ex}{1.0em}\rule[0.5ex]{35.5ex}{0.1em} \tless \rule[0.5ex]{35.5ex}{0.1em}\rule[-0.4ex]{0.2ex}{1.0em}}
    \put(67.7,37){\small \rule[-0.4ex]{0.2ex}{1.0em}\rule[0.5ex]{12.7ex}{0.1em} \lmo \rule[0.5ex]{12.7ex}{0.1em}\rule[-0.4ex]{0.2ex}{1.0em}}
    \end{overpic}\vspace{-8pt}
   \caption{Visual results of 6D object pose estimation by CosyPose (single-view)~\cite{labbe2020cosypose} and StablePose on \tless and \lmo. Please pay special attention to the results of the challenging objects highlighted in the input scene.
   }
   \label{fig:gallery}\vspace{-16pt}
\end{figure*} 

\paragraph{Datasets}
We conduct evaluation on three public datasets: \tless~\cite{hodan2017t}, \lmo~\cite{brachmann2014learning} and \nocs~\cite{wang2019normalized}.
%All the datasets include RGBD images with ground-truth object pose annotations.
\tless is challenging as the objects are texture-less mechanical parts with similar appearance and geometry and most possess symmetries.
\lmo is one of the most widely used datasets for 6D object pose estimation with RGBD images of texture-less household objects with heavy occlusion.
\nocs is used to evaluate category-level object pose estimation. The test set contains unseen objects of five categories with different shape and size from those in the training set.
To better evaluate the generality on unseen objects with larger shape variations, we created a synthetic dataset based on ShapeNet~\cite{chang2015shapenet}, named \snp. This new dataset contains rendered RGBD images of objects from $22$ categories. The test objects are usually significantly different from the training ones in terms of shape and appearance; \supl{please refer to the supplemental material for an overview of this dataset}.

\vspace{-12pt}
\paragraph{Metrics}
%Several options exist for quantitatively evaluating object pose estimation.
For instance-level pose estimation, to facilitate comparison to previous works,
%~\cite{labbe2020cosypose,vidal2018method,drost2010model},
we adopt the 6D localization recall under two popular pose-error functions: Average Closest Point Distance (ADI)~\cite{hinterstoisser2012model} and Visible Surface Discrepancy (VSD)~\cite{hodan2018bop}.
ADI considers both visible and invisible parts of the 3D model surface.
Suppose $\bT$ is the predicted pose and $\bar{\bT}$ the ground-truth, ADI measures the mean Euclidean distance from each point on the 3D model transformed by $\bT$ to the closest point on the model transformed by $\bar{\bT}$.
We use the standard ADI recall metric $e_\text{ADI} < 0.1d$, where $d$ is the object diameter~\cite{hodan2018bop,labbe2020cosypose}.
VSD considers visible parts only.
To evaluate the VSD error of an estimated pose $\bT$ w.r.t. its ground-truth $\bar{\bT}$, we first render the template model into two depth maps $D$ and $\bar{D}$ using the two poses, respectively. VSD measures the differences between $D$ and $\bar{D}$ at image locations where the object is visible in the input observation.
We use the standard VSD recall metric $e_\text{VSD} < 0.3$ with $\tau = 20$mm and $\delta =15$mm~\cite{hodan2018bop,labbe2020cosypose}.
For \tless, the results are reported on 6D localization both for varying number of instances of varying number of objects in single-view RGBD images (VIVO)~\cite{hodan2018bop} and for a single instance of a single object (SISO)~\cite{hodavn2016evaluation}.
For \lmo, since each RGB-D image contains at most one instance for one object, we only report results for SISO.

\subsection{Qualitative Results}
\Fig{gallery} shows some visual results of CosyPose~\cite{labbe2020cosypose} and our StablePose on \tless and \lmo.
These examples encompass objects with heavy occlusion (e.g. column 1--5), discrete symmetry (column 2 and 3) and continuous symmetry (column 4).
In all these challenging cases, our method is able to estimate the 6D object poses accurately.
Especially for those objects with severe occlusion, our poses are quite accurate thanks to the per-patch pose prediction as a reinforcement of the global pose prediction; \emph{see a visual ablation study in the supplemental material}.
%for challenging cases with heavy occlusion, symmetry ambiguity and unseen objects.
%\kxc{make it concrete which ones have these issues and how/why we can do them right.}

\subsection{Quantitative Comparisons}

\paragraph{Comparing to depth-only methods}
We first compare our method to several baselines that take only depth image as input:
%\textbf{Drost et al.}~\cite{drost2010model}: a point pair feature-based method combining a global modeling and a local matching stage.
%\textbf{Vidal et al.}~\cite{vidal2018method}: the state-of-the-art method based on the point pair features.
%\textbf{PointNet++}~\cite{qi2017pointnet++}: a method that takes PointNet++ as the feature extraction backbone and regresses the 6D pose %based on the features. The network is trained by the dense-pixel loss~\cite{wang2019densefusion}.
%\textbf{PointCNN}~\cite{li}: the method replaces the PointNet++ in 3) with a PointCNN.
%\textbf{PPFNet}~\cite{deng2018ppfnet}: a deep learning method that based on intra-patch point pair features. The global feature is aggregated by a max-pooling operation over all the patch features. Object pose is estimated like 3).
\emph{Drost-PPF}~\cite{drost2010model}, \emph{Vidal-PPF}~\cite{vidal2018method}, \emph{PointNet++}~\cite{vidal2018method} and \emph{PPFNet}~\cite{deng2018ppfnet}.
Both \emph{Drost-PPF} and \emph{Vidal-PPF} are state-of-the-art methods based on the point pair features.
They simultaneously detect objects and estimate poses.
We also build two baselines by using PointNet++ and PPFNet as point feature extraction backbone, respectively, followed by a three-layer MLP for regressing 6D pose trained with dense-pixel loss~\cite{wang2019densefusion}.
Like our method, a pretrained object detector~\cite{park2019pix2pose} is use for these methods.
The experiments are conducted on \tless and \lmo; see the results in~\Cref{tab:tless} and ~\Cref{tab:linemod}.
StablePose outperforms all the baselines over both datasets for all metrics.
%PointNet, PointCNN and PPFNet are inferior to our method, due to their preliminary network architectures for 6D pose estimation. %%% This is meaningless.
StablePose outperforms the learning-based methods significantly.
\emph{Drost-PPF} and \emph{Vidal-PPF} are comparably accurate but $3\textapprx 4$ times slower than StablePose inference.
Moreover, the learned StablePose model can generalize to handle unseen objects without a template model (see~\Cref{tab:shapenet}) which is difficult, if not impossible, for non-learning-based methods.
Note, for the experiments on \lmo, we use ADI/ADD, instead of ADI, as \lmo contains many non-symmetric objects~\cite{hodan2018bop}.
%We also notice that our method is more capable of handing object occlusion compared to these two methods (see Figure ??).
%\kxc{Comment is not enough!! Make concrete and interesting discovery!!}

\vspace{-12pt}
\paragraph{Comparing to RGBD methods}
%We also compare to the baselines that work on RGBD images.
%The color images contain rich appearance feature which greatly aids 6D pose estimation.
We compare to the following RGBD baselines: \emph{DenseFusion}~\cite{wang2019densefusion}, \emph{Pix2Pose}~\cite{park2019pix2pose} and \emph{CosyPose} (single-view)~\cite{labbe2020cosypose}.
\emph{DenseFusion} achieves 6D pose estimation by fusing features of RGB and depth.
\emph{Pix2Pose} trains an auto-encoder to regress pixel-wise 3D coordinates.
\emph{CosyPose} is the state-of-the-art method which predicts poses using RGB image followed by an ICP refinement with depth image.
The comparison is conducted on \tless and \lmo; see results in~\Cref{tab:tless} and ~\Cref{tab:linemod}.
%For \lmo, since \emph{DenseFusion} and \emph{CosyPose} did not report the number, we only compare our method to \emph{Pix2Pose}. \kxc{Not a good reason!}
Although using only depth in pose prediction, StablePose beats the RGBD methods on \tless by a large margin.
On \lmo, our method outperforms all methods except \emph{CosyPose} which was trained using a much larger training set of RGB images. The training of StablePose requires a much smaller dataset and hence is significantly faster. This implies that learning on depth/geometric input is more data-efficient for the task of 6D pose estimation.
%A crucial observation of this comparison is that: although CosyPose achieves comparable results to our method, it relies heavily on massive training data and has slower convergence rate (shown in~\Cref{tab:tless}).
%\kxc{Not an observation!! Make concrete and interesting discovery!!}

\begin{table*}
\footnotesize\centering
\caption{
Performance comparison on \tless.
%The table report the recall measured by pose-error functions ADI and VSD on both the VIVO and SISO tasks.
%Our method outperforms all the previous methods in every metric.
%Note that the training data and training time of our method are dramatically smaller than the state-of-the-art RGBD methods~\cite{wang2019densefusion,park2019pix2pose,labbe2020cosypose}, while the inference time of our method outperforms those using hand-crafted features~\cite{drost2010model,vidal2018method} by a large margin.
}\vspace{-8pt}
\label{tab:tless}
%\begin{tabular}{>{\centering}p{60 pt}>{\centering}p{20 pt}>{\centering}p{20 pt}>{\centering}p{20 pt}>{\centering}p{20 pt}>{\centering}p{20 pt}>{\centering}p{20 pt}>{\centering}p{20 pt}p{20 pt}}
\begin{tabular}{ccccccccc}
\toprule
 & Pose Est. & \eadi (VIVO) & \eadi (SISO) & \evsd (VIVO) & \evsd (SISO) &  Training data & Training time & Inference time \\
\midrule
Drost-PPF~\cite{drost2010model} & D & - & - & - & 0.57 & - & - & 1.3s \\
Vidal-PPF~\cite{vidal2018method}              & D & - & - & - & 0.72 & - & - & 1.6s \\
PointNet++~\cite{qi2017pointnet++}         & D & 0.74 & 0.78 & 0.50 & 0.54 & 37K & 15h & 0.4s\\
%PointCNN~\cite{li2018pointcnn}               & D & ? & ? & ? & ? & ? & ? & ?\\
PPFNet~\cite{deng2018ppfnet}               & D & 0.76 & 0.79 & 0.44 & 0.49 & 37K & 15h & 0.4s\\
\midrule
DenseFusion~\cite{wang2019densefusion}               & RGBD & 0.13 & 0.15 & 0.08 & 0.10 & 37K & 15h & 0.1s\\
Pix2Pose~\cite{park2019pix2pose}               & RGBD & - & - & - & 0.30 & 37K & 80h & 0.6s\\
CosyPose~\cite{labbe2020cosypose}               & RGBD & 0.68 & 0.75 & 0.63 & 0.64 & 1M & 200h & 1.1s\\
\midrule
StablePose & D & \textbf{0.86} & \textbf{0.88} & \textbf{0.69} & \textbf{0.73} & 37K & 15h & 0.4s\\
\bottomrule
\end{tabular}
\vspace{1em}\vspace{-15pt}
\end{table*}

\begin{table}
\footnotesize\centering
\caption{
Performance comparison on \lmo.
%Our method significantly outperforms the state-of-the-art methods. This demonstrates its ability to handle object with heavy occlusion.
}\vspace{-8pt}
\label{tab:linemod}
\begin{tabular}{ccccc}
\toprule
 & \eadiadd (SISO) & \evsd (SISO) &  Training data\\ \midrule
Drost-PPF~\cite{drost2010model} & - & 0.55 & - \\
Vidal-PPF~\cite{vidal2018method} & - & 0.62 & - \\
Pix2Pose~\cite{park2019pix2pose} & 0.32 & - & 10K \\
CosyPose~\cite{labbe2020cosypose} & \textbf{0.68} & \textbf{0.83}  & 1M \\
StablePose & 0.63 & 0.71 & 10K \\
\bottomrule
\end{tabular}
\vspace{1em}\vspace{-10pt}
\end{table}

\subsection{Parameter Setting and Ablation Studies}
In~\Cref{tab:ablation}, we study the parameter setting and design choices of our method.

\vspace{-12pt}
\paragraph{Patch count of a stable group}
%We first study whether selecting triplet is the optimal.
Our method selects \emph{three patches} for each stable group.
Here we evaluate other possibilities: \emph{one-patch}, \emph{two-patch} and \emph{five-patch}. All baselines adopt the same network setting as the main method. For one-patch and two-patch cases, the final pose takes average over all groups with uniform weights since stability measure can hardly be computed for less than three patches.
The results show that patch triplet is the best choice for constructing stable groups for pose estimation.
%\kxc{Any more interesting comments?}

\vspace{-12pt}
\paragraph{Patch group sampling}
To evaluate the necessity of \emph{stability-based} patch group sampling, we compare to two baselines: \emph{size-based sampling} and \emph{distance-based sampling}. The former selects patch triplets by point count: Any three patches whose total point count is larger than a threshold ($1000$) form a group. The latter picks patch triplets such that the sum up of pair-wise patch distances exceeds a threshold ($8$ cm).
\Cref{tab:ablation} shows that stability-based sampling performs the best.
This confirms the idea that geometrically stable patches complement better to each other in terms of determining the DoFs of object pose.
%\kxc{Any more interesting comments?}

\vspace{-10pt}
\paragraph{Network design and training scheme}
We also compare to baselines of \emph{without inter-patch contextual feature}, \emph{without patch-wise pose estimation} and \emph{``back-propagate with the best''} in symmetry handling, by ablating each of these algorithmic components.
The degraded performance of the baselines validates our design choices.
Crucially, the performance drops significant for \emph{without patch-wise pose estimation}, which suggests that this auxiliary task indeed provides substantial constraints for improving pose accuracy.

\begin{table}
\footnotesize\centering
\caption{
Parameter setting and ablation studies.
%Ablation study of our method. The results demonstrate that selecting patch groups with three patches outperforms the alternatives.
%Using size-based or distance-based patch group sampling degrades the performance.
%Our full method performs better than the variants that do not include inter-patch feature, point-to-plane/axis loss or the mechanism to handle symmetry.
}\vspace{-8pt}
\label{tab:ablation}
\begin{tabular}{ccc}
\toprule
  & \eadi (SISO) & \evsd (SISO) \\ \midrule
one-patch & 0.70 & 0.57 \\
two-patch & 0.76 & 0.63 \\
five-patch & 0.73 & 0.69 \\  \midrule
%all-patch & 0.67 & 0.72 & ? & ? \\
size-based sampling & 0.85 & 0.69 \\
distance-based sampling & 0.77 & 0.68 \\ \midrule
%w/o inter-patch feature & ? & ? & ? & ? \\
w/o patch-wise pose & 0.73 & 0.62 \\
baseline symmetry handling & 0.75 & 0.54 \\ \midrule
StablePose & \textbf{0.88} & \textbf{0.73} \\
\bottomrule
\end{tabular}
\vspace{1em}\vspace{-15pt}
\end{table}

\subsection{Category-Level Pose Estimation}
The generality of our method can be best reflected by \emph{category-level 6D pose estimation} in which the object instance is unseen during training. We test our method on \nocs and \snp, and compare with two state-of-the-art methods: \emph{NOCS}~\cite{wang2019normalized} and \emph{CASS}~\cite{chen2020learning}.
Note, both \emph{NOCS} and \emph{CASS} predict pose with RGBD, while our method does so with only depth input. The experiments are evaluated using $10^{\circ}10cm$, $IoU25$, $R_{err}$ and $T_{err}$ as in~\cite{wang2019normalized}.
% (\supl{see supplemental material for details}).
In \Cref{tab:shapenet}, we show the results on \snp (\supl{the results on \emph{\nocs} is in the supplemental material}).
%containing more shape variations than \nocs;
The better cross-instance generality of StablePose is due to the repeatability of patch groups across object instances and the pose prediction learning over a redundant set of patch groups.
%\kxc{Interesting comments!!}

%\begin{table}
%\footnotesize\centering
%\begin{tabular}{>{\centering}p{25 pt}>{\centering}p{25 pt}>{\centering}p{25 pt}>{\centering}p{25 pt}>{\centering}p{25 pt}p{25 pt}}
%\toprule
% & Input & $5^{\circ}5cm$ & $IoU25$ & $R_{err}$ & $T_{err}$  \\ \midrule
%NOCS & RGBD & ? & ? & ? & ? \\
%CASS & RGBD & ? & ? & ? & ? \\
%Ours & D & \textbf{0.859} & \textbf{0.884} & \textbf{0.685} & \textbf{0.729} \\
%\bottomrule
%\end{tabular}
%\vspace{1em}
%\caption{
%Comparison of our approach against prior work on category-level pose estimation task. The experiment is conducted on NOCS dataset.
%}
%\label{tab:nocs}
%\end{table}

\begin{table}
\footnotesize\centering
\caption{
Comparison on category-level pose estimation over \snp.
}\vspace{-8pt}
\label{tab:shapenet}
%\begin{tabular}{>{\centering}p{20 pt}>{\centering}p{40 pt}>{\centering}p{20 pt}>{\centering}p{20 pt}>{\centering}p{20 pt}p{20 pt}}
\begin{tabular}{cccccc}
\toprule
 & Pose Est. & $10^{\circ}10cm$ & $IoU25$ & $R_{err}$ & $T_{err}$  \\ \midrule
NOCS~\cite{wang2019normalized} & RGBD & 12.8 & 61.7 & 33.5 & 19.3 \\
CASS~\cite{chen2020learning} & RGBD & 13.9 & 67.3 & 32.9 & 17.6 \\
Ours & D & \textbf{21.4} & \textbf{92.1} & \textbf{20.9} & \textbf{9.6} \\
\bottomrule
\end{tabular}
\vspace{1em}\vspace{-15pt}
\end{table}

%\subsection{Prediction Analysis}
%\ys{plot of coplanarity prediction}
%\ys{analysis by objects} 
%!TEX root = sceneparse.tex

\section{Discussion, Limitations and Future Works}
\label{sec:conclusion}

With our work, we hope to deliver the following key messages.
1) Shape and pose of objects are tightly coupled in visual perception.
2) 6D object pose can be pined down by a minimal set of geometrically stable patches sampled on the object surface.
3) Although each patch determines only a subset of the six DoFs, their under-determined predictions can cummulatively constrain the object pose, resulting in high accuracy and robustness.
We have realized these ideas with StablePose, a multi-task deep neural network with good generalization.
%With a disentangled handling of symmetry ambiguities, the network achieves the state-of-the-art performance on several public benchmarks.
The network achieves the state-of-the-art performance on several public benchmarks.

Our current solution has the following limitations on which future investigations could be conducted.
\emph{First}, object detection still relies on RGB input. Designing a single-stage object detection and pose estimation for depth-only input is an interesting problem to study.
%\emph{Second}, color texture can help resolve some symmetry ambiguities which are unresolvable by geometry.
\emph{Second}, although our method works on redundant set of patch groups, the quality of patch extraction is of great importance. It is difficult to extract valid patches from very small or incompletely scanned objects.
In the future, we plan to realize implicit patch stability analysis in a more end-to-end fashion without explicit patch extraction and stable group sampling.

\section*{Acknowledgements}
We thank the anonymous reviewers for their valuable comments. This work was supported in part by National Key Research and Development Program of China (2018AAA0102200), NSFC (61825305, 62002379) and the Zhejiang Lab’s International Talent Fund for Young Professionals.
%Youth Innovation Project of College of Intelligence Science and Technology, NUDT (2020008).

{\small
\bibliographystyle{ieee_fullname}
\bibliography{egbib}
}

\end{document}

%%%%%%%%% TITLE
%\title{StablePatches: Learning Object Poses from Geometrically Stable Patches\\of a Depth Image}
\title{Supplemental Materials\\StablePose: Learning 6D Object Poses from Geometrically Stable Patches}

\author{
    \hfill
	Yifei Shi\qquad
	Junwen Huang\qquad
	Xin Xu\qquad
    Yifan Zhang\qquad
	Kai Xu
	\hfill
	\vspace{0.1cm}
	\\
	\hfill
	National University of Defense Technology
    \hfill
}

\maketitle

%%%%%%%%% BODY TEXT
%!TEX root = sceneparse.tex

\section{Details of Stability Analysis}
\label{sec:stablity}

Here, we provide the details of stability analysis of point cloud.
Mathematically, given a 3D point set $\mathcal{P}=\{\bv_i, \bn_i\}$ sampled on the template model surface, we want to find a rigid transformation $[\bR|\bt]$ which minimizes the following point-to-plane error at all points:
\begin{equation}\label{eq:alignerror}
\min_{[\bR,\bt]}\sum_{i}{[(\bR\bv_i+\bt)\cdot \mathbf{n}_i]^2},
\end{equation}
where $\bR$ and $\bt$ are rotation and translation, respectively.

The rotation $\bR$ is nonlinear but can be linearized assuming infinitesimal rotations:
\begin{equation}\label{eq:rotlinear}
\mathbf{R} \approx \begin{pmatrix}
                     1 & -\gamma & \beta \\
                     \gamma & 1 & -\alpha \\
                     -\beta & \alpha & 1 \\
                   \end{pmatrix},
\end{equation}
for Euler angles $\alpha$, $\beta$, and $\gamma$ around the X, Y, and Z axes, respectively.
This reduces the rotation of $\bv_i \in \mathbf{V}$ by $\bR$ into a
displacement of it by a vector $[\br \times \bv_i + \bt]$, where
$\br=(\alpha, \beta, \gamma)$. Substituting this into Equation (\ref{eq:alignerror}),
we therefore aim to find a 6-vector $[\br^T, \bt^T]$ that minimizes
\begin{equation}\label{eq:linearerror}
\min_{[\br,\bt]}\sum_{i}{[\bv_i \cdot \bn_i + \br \cdot (\bv_i \times \bn_i) + \bt \cdot \bn_i]}.
\end{equation}
%We minimize $\mathcal{E}$ with respect to $\alpha$, $\beta$, $\gamma$, $\mathbf{t}_x$, $\mathbf{t}_y$, and $\mathbf{t}_z$ by setting their partial derivatives to zero:
%\begin{equation}\label{eq:derivatives}
%\begin{aligned}
%\frac{\partial \mathcal{E}}{\partial \alpha}= \sum_{i}2c_{i,x}[\mathbf{p}_i \cdot \mathbf{n}_i + \mathbf{r} \cdot (\mathbf{p}_i \times \mathbf{n}_i) + \mathbf{t} \cdot \mathbf{n}_i]=0\\
%\frac{\partial \mathcal{E}}{\partial \beta}= \sum_{i}2c_{i,y}[\mathbf{p}_i \cdot \mathbf{n}_i + \mathbf{r} \cdot (\mathbf{p}_i \times \mathbf{n}_i) + \mathbf{t} \cdot \mathbf{n}_i]=0\\
%\frac{\partial \mathcal{E}}{\partial \gamma}= \sum_{i}2c_{i,z}[\mathbf{p}_i \cdot \mathbf{n}_i + \mathbf{r} \cdot (\mathbf{p}_i \times \mathbf{n}_i) + \mathbf{t} \cdot \mathbf{n}_i]=0\\
%\frac{\partial \mathcal{E}}{\partial t_x}= \sum_{i}2n_{i,x}[\mathbf{p}_i \cdot \mathbf{n}_i + \mathbf{r} \cdot (\mathbf{p}_i \times \mathbf{n}_i) + \mathbf{t} \cdot \mathbf{n}_i]=0\\
%\frac{\partial \mathcal{E}}{\partial t_y}= \sum_{i}2n_{i,y}[\mathbf{p}_i \cdot \mathbf{n}_i + \mathbf{r} \cdot (\mathbf{p}_i \times \mathbf{n}_i) + \mathbf{t} \cdot \mathbf{n}_i]=0\\
%\frac{\partial \mathcal{E}}{\partial t_z}= \sum_{i}2n_{i,z}[\mathbf{p}_i \cdot \mathbf{n}_i + \mathbf{r} \cdot (\mathbf{p}_i \times \mathbf{n}_i) + \mathbf{t} \cdot \mathbf{n}_i]=0\\
%\end{aligned}
%\end{equation}
%
This is a linear least-squares problem which amounts to solve a linear system $C\mathbf{x}=0$ with $\mathbf{x}=[\br^T, \bt^T]$. $C$ is a $6\times 6$ covariance matrix of the rigid transformation accumulated over all sample points:
\begin{equation}\label{eq:covariance}
C=\sum_{i}\left[
             \begin{array}{cccccc}
             \mathbf{u}_{ix} \\ \mathbf{u}_{iy} \\ \mathbf{u}_{iz} \\ \mathbf{n}_{ix} \\ \mathbf{n}_{iy} \\ \mathbf{n}_{iz} \\
             \end{array}
\right]
\left[
             \begin{array}{c}
             \mathbf{u}_{ix} \ \mathbf{u}_{iy} \  \mathbf{u}_{iz} \  \mathbf{n}_{ix} \  \mathbf{n}_{iy} \  \mathbf{n}_{iz} \\
             \end{array}
\right],
\end{equation}
%\begin{equation}\label{eq:covariance}
%\sum_{i}\left[
%             \begin{array}{cccccc}
%             \mathbf{c}_{ix}\mathbf{c}_{ix} & \mathbf{c}_{ix}\mathbf{c}_{iy} & \mathbf{c}_{ix}\mathbf{c}_{iz} & \mathbf{c}_{ix}\mathbf{n}_{ix} & \mathbf{c}_{ix}\mathbf{n}_{iy} & \mathbf{c}_{ix}\mathbf{n}_{iz} \\
%             \mathbf{c}_{iy}\mathbf{c}_{ix} & \mathbf{c}_{iy}\mathbf{c}_{iy} & \mathbf{c}_{iy}\mathbf{c}_{iz} & \mathbf{c}_{iy}\mathbf{n}_{ix} & \mathbf{c}_{iy}\mathbf{n}_{iy} & \mathbf{c}_{iy}\mathbf{n}_{iz}\\
%             \mathbf{c}_{iz}\mathbf{c}_{ix} & \mathbf{c}_{iz}\mathbf{c}_{iy} & \mathbf{c}_{iz}\mathbf{c}_{iz} & \mathbf{c}_{iz}\mathbf{n}_{ix} & \mathbf{c}_{iz}\mathbf{n}_{iy} & \mathbf{c}_{iz}\mathbf{n}_{iz}\\
%             \mathbf{n}_{ix}\mathbf{c}_{ix} & \mathbf{n}_{ix}\mathbf{c}_{iy} & \mathbf{n}_{ix}\mathbf{c}_{iz} & \mathbf{n}_{ix}\mathbf{n}_{ix} & \mathbf{n}_{ix}\mathbf{n}_{iy} & \mathbf{n}_{ix}\mathbf{n}_{iz}\\
%             \mathbf{n}_{iy}\mathbf{c}_{ix} & \mathbf{n}_{iy}\mathbf{c}_{iy} & \mathbf{n}_{iy}\mathbf{c}_{iz} & \mathbf{n}_{iy}\mathbf{n}_{ix} & \mathbf{n}_{iy}\mathbf{n}_{iy} & \mathbf{n}_{iy}\mathbf{n}_{iz}\\
%             \mathbf{n}_{iz}\mathbf{c}_{ix} & \mathbf{n}_{iz}\mathbf{c}_{iy} & \mathbf{n}_{iz}\mathbf{c}_{iz} & \mathbf{n}_{iz}\mathbf{n}_{ix} & \mathbf{n}_{iz}\mathbf{n}_{iy} & \mathbf{n}_{iz}\mathbf{n}_{iz}\\
%             \end{array}
%\right]
%\end{equation}
where $\mathbf{u}=\mathbf{v}\times \mathbf{n}$.
The covariance matrix $C$ encodes the increase of the point-to-plane error when the transformation is moved away from its optimum. The larger the error increase, the less slippable and  more stable along that transformation the shape is. On the contrary, if there is a transformation that causes small increase in the error, the shape is unstable w.r.t. the corresponding DoFs.

%By expanding $C$ in terms of its eigenvectors we may see directly the effect of various incremental transformations.
The stability can then be analyzed by calculating the eigenvalues of $C$.
Let $\lambda_1 \leq \lambda_2 \leq \cdots \leq \lambda_6$ be the eigenvalues of $C$.
%If some eigenvalue $\lambda_j$ is small (i.e., $\frac{\lambda_6}{\lambda_j}$ is greater than a threshold), its corresponding eigenvector (transformation) causes small error increase, meaning that the shape is unstable w.r.t the corresponding DoFs.
The stability is measured as $[1+e^{0.05\left(\frac{\lambda_6}{\lambda_1}-200\right)}]^{-1}$, where $\lambda_1$ and $\lambda_6$ are the smallest and largest eigenvalues of $C$, respectively.
%Otherwise, if no eigenvalue is small, the shape is stable under rigid transformation.
In our method, we select patch group whose stability measure is greater than $0.5$ as the geometrically stable patch group.

{\small
\bibliographystyle{ieee_fullname}
\bibliography{egbib}
}